\pdfoutput=1

\documentclass[11pt]{article}

\usepackage[]{EACL2023}

\usepackage{times}
\usepackage{latexsym}
\usepackage{multirow}
\usepackage{multicol}
\usepackage{booktabs}
\usepackage{amsmath}
\usepackage{pgfplots}
\pgfplotsset{compat=1.18}

\usepackage[T1]{fontenc}

\usepackage[utf8]{inputenc}

\usepackage{microtype}

\usepackage{inconsolata}

\title{Adapting the NICT-JLE Corpus for Disfluency Detection Models}

\author{Lucy Skidmore \\
  University of Sheffield, UK \\
  \texttt{lskidmore1@shef.ac.uk} \\\And
  Roger K. Moore \\
  University of Sheffield, UK \\
  \texttt{r.k.moore@shef.ac.uk} \\}

\begin{document}
\maketitle
\begin{abstract}
The detection of disfluencies such as hesitations, repetitions and false starts commonly found in speech is a widely studied area of research. With a standardised process for evaluation using the Switchboard Corpus, model performance can be easily compared across approaches. This is not the case for disfluency detection research on learner speech, however, where such datasets have restricted access policies, making comparison and subsequent development of improved models more challenging. To address this issue, this paper describes the adaptation of the NICT-JLE corpus, containing approximately 300 hours of English learners' oral proficiency tests, to a format that is suitable for disfluency detection model training and evaluation. Points of difference between the NICT-JLE and Switchboard corpora are explored, followed by a detailed overview of adaptations to the tag set and meta-features of the NICT-JLE corpus. The result of this work provides a standardised train, heldout and test set for use in future research on disfluency detection for learner speech.   

\end{abstract}

\section{Introduction}

Consider the utterance below: 
\begin{align*}
{\text{I'd like a }\big[ \underbrace{\text{coffee}}_\textrm{reparandum} \text{\large+} \underbrace{\text{\{uh\}}}_\textrm{interregnum} \underbrace{\text{tea}}_\textrm{repair} \big] \text{ please}}
\end{align*}

The speaker initially asks for a coffee but changes their request to tea instead. The linguistic mechanism by which this is achieved is known as self-repair, or in computer science research, disfluency. Disfluencies are comprised of a reparandum phrase, optional interregnum phrase and repair phrase, often marked prosodically at the `interruption point' + with features such as silence or reparandum word cutoff \cite{levelt1983monitoring}. Interregna can contain filled pauses such as ``\textit{uh}'' like in the example, edit terms such as ``\textit{I mean}'' and finally discourse markers such as ``\textit{you know}''.

There are two perspectives from which disfluency detection research is applied. The first is disfluency removal, where transcribed speech is transformed into a form more similar to written text for subsequent downstream NLP tasks. The second is incremental detection, whereby disfluent phrases are detected word-by-word and retained to infer meaning for use in spoken dialogue systems. The most successful approaches leverage BERT language models \citep{devlin-etal-2019-bert} to achieve high accuracy for both non-incremental (\citealp{bach19_interspeech, jamshid-lou-johnson-2020-improving, rocholl21_interspeech}) and incremental \citep{rohanian-hough-2021-best} frameworks. 

Disfluency detection has also been explored for learner speech, using parsing-based approaches \citep{moore2015incremental}, bi-directional LSTMs \citep{lu19_slate} and end-to-end models \citep{lu20e_interspeech} for the downstream task of grammatical error correction. These approaches train models using native speech and evaluate using datasets with restricted access  \citep{BULATS_annotation} or small subsets of publicly available data \citep{NICT}. This approach is undesirable for two reasons: \textit{(i)} model performance cannot easily be compared with other work due to the limited access to evaluation materials for the task, and \textit{(ii)}, models trained on native data creates a performance bias towards higher proficiency learners \citep{moore2015incremental}.   

With the above in mind, this work expands on that of  \citet{skidmore-moore-2022-incremental}, who trained and evaluated an incremental disfluency detection model using the NICT-JLE corpus. The corpus is introduced and its features are explored through a comparative analysis with the Switchboard corpus. This is followed by a detailed overview of the tag set adaptation, POS-tagging approach, additional meta-features and the train, heldout and test divisions. This paper concludes with a discussion of directions for future work as well as limitations of the adapted corpus.


\section{The NICT-JLE Corpus}

The National Institute of Information and Communications Technology Japanese Learner English (NICT-JLE) Corpus contains approximately 300 hours of transcribed oral proficiency tests of 1,281 Japanese-speaking learners of English \citep{NICT}. The Standard Speaking Test (SST) is made up of three tasks for a learner to carry out with the assessor: open dialogue, a role-play scenario and a picture description task. Each transcribed test contains HTML-style tags for edit terms and disfluencies, `non-verbal sounds' (including silence and laughter), as well as meta-data such as the learners' proficiency level, gender and nationality. 

\subsection{Disfluencies in the NICT-JLE corpus}

\setlength{\tabcolsep}{4.5pt}
\begin{table}[t]
\centering
\caption{\label{tab:linguistic} General linguistic and disfluency features of the NICT-JLE (NICT) and Switchboard (SWBD) corpora.}
{
\begin{tabular}{rll}
\toprule
 & NICT & SWBD \\ 
\midrule
total words                     & 1165785       & 746290 \\
total utterances                & 178934        & 102169 \\
vocabulary size                 & 13499         & 16810 \\
utterance length (SD)   & 6.51 (3.27)   & 7.30 (3.61) \\
\midrule
disfluency/100 words & 7.54 & 3.56 \\
edit term/100 words & 11.55 & 5.16 \\
rm length (SD) & 1.62 (1.08) & 1.58 (1.12) \\
nested rate & 39.35 & 21.68 \\
non-repetitious rate & 45.60 & 49.45 \\
with-interregnum rate & 31.09 & 22.17 \\ 
\bottomrule
\end{tabular}}
\end{table}

Table \ref{tab:linguistic} compares a range of general linguistic and disfluency features of the NICT-JLE corpus with the Switchboard corpus---the standard corpus used for disfluency detection on native speech. The NICT-JLE corpus is the larger of the two, with a higher number of both words and utterances. The Switchboard corpus has a larger vocabulary size and longer average utterance length. These figures are reflective of second language acquisition research that determines both vocabulary size and average utterance length as predictors of learners' speaking skills \cite{koizumi2013vocabulary, hilton2008link}, where a larger vocabulary equates to a higher speaking proficiency. 

Looking at the disfluency features, the NICT-JLE corpus has over twice as many disfluencies and edit terms per 100 words compared to the Switchboard corpus. These figures are again echoed in prior research, which attributes the lower degree of language `automatisation' of language learners to the increased number of disfluencies found in learner speech \cite{wiese1984language, temple1992disfluencies}, with the same behaviour also seen for filled pauses \cite{hilton2008link, de2013linguistic}. Reparandum phrase lengths, however, are comparable between the two corpora. Finally, when considering the features of the disfluencies themselves, the NICT-JLE corpus has higher rates of both nested and with-interregnum disfluencies, whereas the Switchboard corpus has a higher rate of non-repetitious disfluencies.

\begin{figure}[t]
  \centering
\begin{tikzpicture}
  \pgfkeys{
    /pgf/number format/precision=0,
    /pgf/number format/fixed zerofill=true,
    /pgf/number format/fixed
  }
  \begin{axis}[%
      ybar,
      width=0.5\textwidth,
      height=6.5cm,
      bar width=0.15cm,
      ymax=11,
      ymin=0,
      ytick = {0, 2, 4, 6, 8, 10},
      enlarge x limits={0.2, auto},
      clip=false,
      legend pos= {north east},
      legend cell align={left},
      ylabel={frequency per 100 words},
      xlabel={speaker proficiency level}, 
      symbolic x coords={beginner,intermediate,advanced,native},
      xtick=data,
      tick label style={font=\small},
      legend style={font=\small}
    ]
        
    \addplot [fill=black!20!white, area legend, 
    error bars/.cd, y dir=both, y explicit
        ] coordinates {
            (beginner,9.83)
            (intermediate,8.02) 
            (advanced,4.51) 
            (native,3.56)
        };
        
    \addplot [fill=teal!40!white, area legend, 
    error bars/.cd, y dir=both, y explicit
        ] coordinates {
            (beginner,3.75)
            (intermediate,3.61) 
            (advanced,2.63)
            (native,1.76)
        };   
        
    \addplot [fill=yellow!20!white, area legend, 
    error bars/.cd, y dir=both, y explicit
        ] coordinates {
            (beginner,6.08)
            (intermediate,4.41) 
            (advanced,1.89)
            (native, 1.80)
        };  
        
    \addplot [fill=teal, area legend, 
    error bars/.cd, y dir=both, y explicit
        ] coordinates {
            (beginner,2.66)
            (intermediate,2.25) 
            (advanced,0.97)
            (native,0.79)
        };
        
    \addplot [fill=red, area legend, 
    error bars/.cd, y dir=both, y explicit
        ] coordinates {
            (beginner,4.65)
            (intermediate,3.14) 
            (advanced,1.37)
            (native,0.77)
        };
        
    \legend{all instances, non-repetitious, repetitious, with-interregnum, nested}
  \end{axis}
\end{tikzpicture}
  \caption{Frequency per 100 words of disfluency types according to speaker proficiency level.}
  \label{fig:levels}
\end{figure}
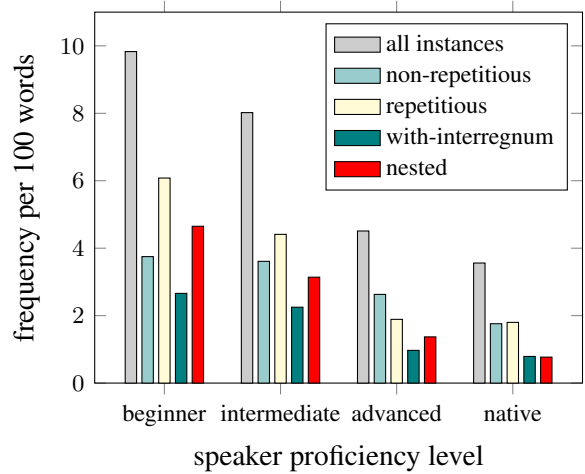

Exploring the relationship between disfluency behaviour and proficiency level further, Figure \ref{fig:levels} shows the frequency per 100 words of all disfluency instances, and subsequently, repetitious, non-repetitious, with-interregnum and nested instances according to speaker proficiency level. The `beginner', `intermediate', and `advanced' levels refer to speakers from the NICT-JLE corpus at SST levels one to three, four to six and seven to nine, respectively. The `native' group refers to all speakers in the Switchboard corpus. As can be seen, the frequency of all disfluency types decreases as speaker proficiency level increases. Notably, the ratio of repetitious to non-repetitious disfluency instances is at its highest for beginner proficiencies and marginally higher for intermediate proficiencies. For advanced speakers, the distribution switches and non-repetitious disfluencies are shown to be more frequent. For native speakers, the distributions are approximately equal. These observations are again reflective of prior research, which has shown the frequency of repetitions in learner speech to decline as linguistic knowledge increases \cite{de2013linguistic} as well as the distribution of features becoming more similar to that of a native speaker \cite{van1996self}. 

\subsection{Learner errors in the NICT-JLE corpus}

The NICT-JLE corpus also contains learner errors both inside and outside of disfluencies. The examples below illustrate instances of the former, with errors occurring in the reparandum phrase, the repair phrase, or both. The disfluency phrases are labelled and words in bold indicate learner errors.  

\medskip 

\begin{center}
  (1) My computer [\textbf{use} + \{er\} is used] by [\textbf{all family} + my family]
\end{center}

\begin{center}
  (2) She [[\textbf{wanted shopping} + \textbf{wanted shop}] + \{er\} wanted to go shopping]
\end{center}

\begin{center}
  (3) [[I don't + \textbf{I'm not have watching movie}] + I don't have \textbf{no} time to \textbf{watch movie}] 
\end{center}

\medskip 

There are 167 interviews in the NICT-JLE corpus that contain additional labels for learners' morphological, grammatical and lexical errors. From this subset, there are approximately 11 instances of learner errors per 100 words. However, the actual rate of errors in the corpus is likely to be higher as errors that occur as part of a reparandum phrase are not annotated. 


\section{Adapting the NICT-JLE Corpus}

The adapted version of the NICT-JLE corpus is available online\footnote{https://github.com/lucyskidmore/nict-jle}. With the focus of this task being learner speech, any utterances of assessors are omitted. In addition, utterances that contain Japanese, or partial sentences due to the retraction of personally identifiable details are removed. Individual file numbers are retained and utterance segmentation follows the original transcriptions.

\begin{figure*}
    \centering
    \includegraphics[width=0.95\textwidth]{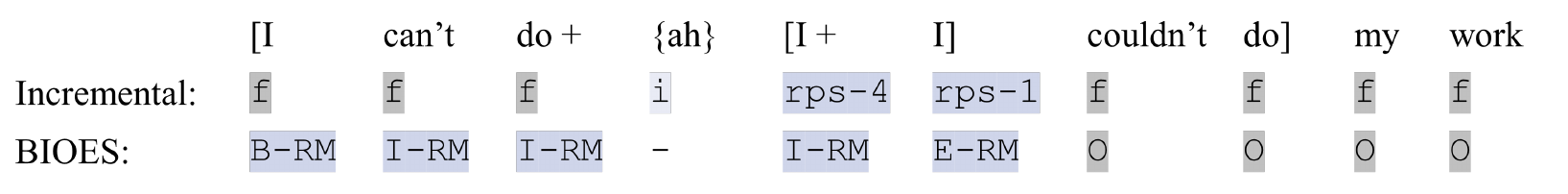}
    \caption{The incremental and BIOES tag sets for the adapted NICT-JLE corpus.}
    \label{fig:tags}
\end{figure*}

\subsection{Tokenization and POS tags}

The corpus was tokenized using the Natural Language Toolkit (NLTK) \citep{bird2009natural} and subsequently tagged with parts-of-speech (POS) using Stanford's left3words MaxentTagger \cite{toutanova2003feature}. Following the same conventions as the Switchboard corpus, contracted forms of words such as ``I've'' and ``can't'' were split by tokenization and re-merged after POS tag labelling to form compounded POS tags. In the case of the two examples, the POS tags for these words are \texttt{PRPVBP} and \texttt{MDRB}, respectively. 

\subsection{Disfluency tag sets}


In order to match the conventions used in previous approaches to disfluency detection, the original disfluency labels in the NICT-JLE corpus were adapted to include two labelling styles: incremental tags \citep{hough15_interspeech} and ‘beginning-inside-outside-end-single’ (BIOES) tags \citep{lu19_slate, lu20e_interspeech}, depicted in Figure \ref{fig:tags}. As can be seen, the incremental approach includes the repair phrase start in the tag set, with the relative position of the reparandum start denoted by \texttt{rps-n}. Here \texttt{f} refers to `fluent' text and \texttt{i} refers to interregna. The BIOES tags only label the reparandum phrase, as `single reparandum' \texttt{S-RM} for one word reparanda and as `begin reparandum' \texttt{B-RM}, `inside reparandum' \texttt{I-RM} or `end of reparandum' \texttt{E-RM} for longer reparandum phrases. All other tokens are considered `outside' \texttt{O} of the disfluent phrase. This approach removes the nesting structure from disfluencies to create one large disfluency instance. Interregna tags are omitted from the BIOES approach as they are often omitted from the tag set altogether during experimentation \citep{lu19_slate, lu20e_interspeech}. The original labelling scheme for the NICT-JLE corpus does not include repair phrases so they are not included in the adapted tag sets.   

\subsection{Meta features}

Alongside labels for disfluencies and learner errors, the NICT-JLE corpus also contains transcriptions for learners' `non-verbal sounds'. From these annotations, silence and laughter were included as features in the adapted dataset due to their prevalence in the corpus as well as their value in previous disfluency detection \citep{ferguson2015disfluency, lu20e_interspeech} and dialogue processing \citep{maraev2021dialogue} applications. In the adapted corpus, each word is additionally assigned four binary values (1 or 0), indicating the presence or absence of a preceding short pause, long pause and laughter, as well as if the word itself was laughed. For the test set only, an additional binary value for each word is included to indicate the presence of a learner error. For `omittance' type errors such as missed prepositions, the word preceding the omittance is labelled as erroneous. For example, in the utterance ``I eat dinner in centre of Tokyo'', `in' would be labelled. Information regarding task type is also retained, due to the proven influence of activity on leaner disfluency behaviour \citep{kormos1998new}. Each word is labelled as `conversation', `picture description' or `role play', relating to the current activity type of the learner. For the latter two, the corpus also includes information regarding the topic of the activity. Finally, the speaker-level features of gender and English proficiency level (SST score) are also included.

\subsection{Train, heldout and test datasets}

\setlength{\tabcolsep}{4.5pt}
\begin{table}[t]
\caption{\label{tab:final_stats} General disfluency features for the train, heldout and test sets of the adapted NICT-JLE corpus.}
\centering
{
\begin{tabular}{rlll}
\toprule
& Train & Heldout & Test \\ 
\midrule
disfluency/100 words & 7.53  & 7.20  & 7.89  \\
edit term/100 words & 11.53  & 11.30 & 11.90  \\
rm length & 2.04  & 2.02  & 2.08 \\

\bottomrule
\end{tabular}}
\end{table} 

Following the approach of the Switchboard corpus, the adapted NICT-JLE corpus was split with 80\% of the files for training, 10\% for heldout and 10\% for testing. Each set has a near-equal distribution of learners according to English proficiency and all data for the test set has accompanying learner error tags. Table \ref{tab:final_stats} shows the disfluency statistics for each of the split datasets, showing equivalency across sets.    




\section{Discussion}

The comparative analysis of the NICT-JLE and Switchboard corpora reveals key differences in terms of linguistic complexity and disfluency behaviour. The native speech of the Switchboard corpus is more lexically complex, with longer utterances, wider vocabulary and a lower ratio of repetitious disfluencies. The disfluencies of the NICT-JLE corpus can be considered more `stutter-like' with higher rates of edit terms, repetitions and learner errors. The parallelism between the data reported here and prior research on learner disfluency behaviour provides support for the adapted NICT-JLE corpus to be used as a proxy for learner speech more generally, highlighting key areas of attention for future research. Examples of such areas include the impact of nested disfluencies, edit terms and learner errors on detection accuracy. 

The inclusion of various tag sets for incremental and non-incremental approaches provides ample opportunity for future research using the adapted NICT-JLE corpus. The most pertinent of which would be to apply the current state-of-the-art approaches using BERT language models \citep{bach19_interspeech, rohanian-hough-2021-best} to a language learning setting, using the adapted NICT-JLE corpus for fine-tuning. Furthermore, the inclusion of prosodic and speaker-level features not only allows for multimodal models to be developed, such as those introduced by \citet{skidmore-moore-2022-incremental}, but also provides a framework for more in-depth model analysis from a language learning perspective, such as the impact of activity type and speaking proficiency on model performance.   Finally, the adapted corpus has the potential for further development. One example would be the inclusion of language assessors' utterances in order to accommodate other dialogue-processing tasks such as end-of-turn prediction and utterance segmentation.

\section{Summary}

In summary, this work details the adaptation of the NICT-JLE corpus to be used as training and evaluation data for disfluency detection in learner speech. A comparative analysis of the NICT-JLE and Switchboard corpora confirmed the NICT-JLE corpus to be a suitable proxy for learner speech and identified multiple challenges for future disfluency detection models to address. The standardised train, heldout and test sets developed here not only allow for a fair comparison between any future models that are developed but also provide a framework for expanding model evaluation to areas that are important for language learning applications. 

\section*{Limitations}

The main limitations of this work are due to the original labelling approach taken for the development of the NICT-JLE corpus, the first of which is the labelling of learner errors. With only a subset of files labelled for errors, together with errors that occur in the reparandum going unlabelled, there is a limited scope for model development or analysis using these features. As explored above, errors can occur as part of disfluencies, and having a full dataset labelled with these occurrences would be valuable, not only for evaluation but also for joint modelling of the linguistic phenomena. The second limitation is that the approach to disfluency labelling is not directly comparable to that of the Switchboard corpus, as the NICT-JLE corpus does not contain labels for repair phrases and follows different nesting rules than those set out by \citet{shriberg1994preliminaries}.     

The NICT-JLE corpus is additionally limited in that it only contains speech from Japanese-speaking learners of English. With the knowledge that learners' native language can influence factors such as disfluency frequency \cite{zuniga2019factors}, it would be valuable to collect data from learners with varied first language backgrounds. In a similar vein, using a POS tagger specifically developed for learner speech such as that described by \citet{nagata-etal-2018-pos} would be beneficial for future iterations of the adapted corpus.

\section*{Acknowledgements}
Funding for this work was awarded through the University of Sheffield Publication Scholarship scheme. 

\bibliography{anthology,custom}
\bibliographystyle{acl_natbib}




\end{document}